\documentclass[11pt]{article}
\usepackage{acl2016}
\usepackage{times}
\usepackage{latexsym}
\usepackage{url}
\usepackage{arydshln}
\usepackage{graphicx}
\usepackage{amsmath}
\usepackage{amsfonts, amssymb}
\usepackage{xcolor}

\newcommand{\todoon}{\long\gdef\todo##1{{
\bf\textcolor{red} {TODO: ##1}
}}}
\todoon

\newcounter{notecounter}
\newcommand{\enotesoff}{\long\gdef\enote##1##2{}}
\newcommand{\enoteson}{\long\gdef\enote##1##2{{
\stepcounter{notecounter}
{\large\bf
\hspace{1cm}\arabic{notecounter} $<<<$ ##1: ##2
$>>>$\hspace{1cm}}}}}
\enoteson
\enotesoff

\def\uprmdn#1#2{\mbox{$_{\scriptsize #2}^{\hbox{\scriptsize #1}}$}}
\def\uprmmath#1#2{_#2^{#1}}

\def\figref#1{Figure~\ref{fig:#1}}
\def\figlabel#1{\label{fig:#1}}

\def\tabref#1{Table~\ref{tab:#1}}
\def\tablabel#1{\label{tab:#1}\label{p:#1}}

\def\secref#1{Section~\ref{sec:#1}}
\def\seclabel#1{\label{sec:#1}\label{p:#1}}
\def\eqref#1{Eq.~\ref{eqn:#1}}

\aclfinalcopy 
\def\aclpaperid{93} 

\title{Intrinsic Subspace Evaluation of Word Embedding Representations }

\author{Yadollah Yaghoobzadeh \rm{and} \textbf{Hinrich Sch{\"u}tze}\\
  Center for Information and Language Processing \\
  University of Munich, Germany \\
  {\tt yadollah@cis.lmu.de} 
}
\begin{document}
\maketitle
\begin{abstract}
We introduce a new methodology for intrinsic evaluation of
word representations.  Specifically, we identify four
fundamental criteria based on the characteristics of natural
language that pose difficulties to NLP systems; and develop
tests that directly show whether or not representations contain \emph{the
subspaces necessary to satisfy these criteria}.
Current intrinsic evaluations are mostly based on the
overall similarity or \emph{full-space similarity}
of words and thus view vector representations as
\emph{points}.
We
show the limits of these point-based intrinsic evaluations.
We apply
our evaluation methodology to the comparison of a count
vector model and several neural network models and demonstrate
important properties of these models.
\end{abstract}


\section{Introduction}
Distributional word representations or \emph{embeddings}
are
currently an
active area of research
in natural language processing (NLP).
The motivation for embeddings is that
knowledge about words is helpful in NLP.  Representing words
as vocabulary indexes may be a good approach if large
training sets allow us to learn everything we need to know
about a word to solve a particular task; but in most cases
it helps to have a representation that contains
distributional information and allows inferences like:
``above'' and ``below'' have similar syntactic behavior or
``engine'' and ``motor'' have similar meaning.

Several methods have been introduced to assess the quality
of word embeddings. We distinguish two different types of
evaluation in this paper: (i) \emph{extrinsic evaluation} evaluates
embeddings in an NLP application or task and (ii) \emph{intrinsic
evaluation} tests the quality of representations independent
of a specific NLP task.

Each single word 
is a combination of a large number of morphological, lexical, syntactic,
semantic, discourse and other features. 
Its embedding should accurately and consistently represent these features, 
and ideally a good evaluation method must clarify this and give
a way to analyze the results.
The goal of this paper is to build such an evaluation.

Extrinsic evaluation is a valid methodology, but it does
not allow us to understand the properties of 
representations without further analysis; e.g.,
if an evaluation shows that embedding A works better than
embedding B on a task, then that is not an analysis of
the causes of the improvement.
Therefore, extrinsic evaluations do not satisfy our goals.

Intrinsic evaluation analyzes the generic quality of embeddings.
Currently, this evaluation mostly is done
by testing overall distance/similarity of words in the embedding space, i.e.,
it is based on viewing word representations as points and
then computing \emph{full-space similarity}.
The assumption is that the high dimensional space is smooth
and similar words are close to each other.  
Several datasets have been developed 
for this purpose,
mostly the result of human judgement;
see \cite{baroni14predict} for an overview.
We refer to these evaluations as \emph{point-based}  and as \emph{full-space}
because they consider embeddings 
as points in the space -- sub-similarities in subspaces are generally ignored.


Point-based intrinsic evaluation computes a score based on
the full-space similarity of two words: a single number that
generally does not say anything about the underlying reasons
for a lower or higher value of full-space similarity.  This
makes it hard to interpret the results of point-based
evaluation and may be the reason that contradictory results
have been published; e.g., based on point-based evaluation,
some papers have claimed that count-based representations
perform as well as learning-based representations
\cite{levy14linguistic}. Others have claimed the opposite
(e.g., \newcite{mikolov2013efficient},
\newcite{pennington14glove}, \newcite{baroni14predict}).

Given the limits of current evaluations, we propose a new
methodology for intrinsic evaluation of embeddings by
identifying generic fundamental criteria for embedding
models that are important for representing features of words
accurately and consistently.  We develop corpus-based tests
using supervised classification that directly show whether
the representations contain the information necessary to
meet the criteria or not.  The fine-grained corpus-based
supervision makes the sub-similarities of words important by
looking at the subspaces of word embeddings relevant to the
criteria, and this enables us to give direct insights into
properties of representation models.


\section{Related Work}
\newcite{baroni14predict} evaluate embeddings on different intrinsic tests:
similarity, analogy, synonym detection, categorization and selectional preference.
\newcite{schnabel15evals} introduce 
tasks
with more fine-grained datasets.
These tasks are unsupervised and 
generally based on cosine similarity; this means that only the
overall direction of vectors is considered or, equivalently,
that \emph{words are modeled as points} in a space and only their
full-space distance/closeness is considered.
In contrast, we test embeddings in a classification
setting and \emph{different subspaces of embeddings are analyzed}.
\newcite{tsvetkov15eval} evaluate embeddings based on their 
correlations with WordNet-based linguistic  embeddings.
However, correlation does not directly evaluate how accurately and completely an application can extract a particular piece of information from an embedding.

Extrinsic evaluations are also common (cf.\ \cite{LiJ15a,koehn15,lai2015compoare}).
\newcite{LiJ15a} conclude that 
embedding evaluation must go beyond human-judgement tasks like similarity and analogy. 
They suggest to evaluate on NLP tasks.
\newcite{koehn15} gives similar suggestions and also recommends
the use of supervised methods for evaluation.
\newcite{lai2015compoare} evaluate embeddings in different tasks
with different setups and show the contradictory results
of embedding models on different tasks.
Idiosyncrasies of different downstream tasks can affect
extrinsic evaluations and result in 
contradictions.

\section{Criteria for word representations}
Each word is a combination of different properties. 
Depending on the language, these properties include
  lexical, syntactic, semantic,
  world knowledge and other features. 
We call these properties \emph{facets}.
The ultimate goal is to learn representations for words that accurately and consistently 
contain these facets. 
Take the facet gender (GEN) as an example.  We call a
representation 100\% \emph{accurate} for GEN if information
it contains about GEN is always accurate; we call the
representation 100\% \emph{consistent} for GEN if the
representation of every word that has a GEN facet contains
this information. 

We now introduce four important
criteria that a representation must satisfy 
to represent facets accurately and consistently. 
These criteria are applied across different problems that
NLP applications face in the effective use of embeddings.

\textbf{Nonconflation.}
A word embedding must 
keep
the evidence from different local contexts
separate -- ``do not conflate'' --
because each context can
infer specific facets of the word. 
Embeddings for different word forms with the same stem, 
like plural and singular forms or different 
verb tenses, are examples vulnerable to conflation
because they occur in similar contexts.

\textbf{Robustness against sparseness.}
One aspect of natural language that poses great
difficulty for statistical modeling is sparseness. 
Rare words are common
in natural language
and embedding models must learn useful 
representations based on a small number of contexts. 

\textbf{Robustness against ambiguity.}
Another central problem when processing words in NLP is
lexical ambiguity \cite{cruse86lexical,zhong2010makes}.
Polysemy and homonymy of words can make it difficult for
a statistical approach to generalize and infer well. 
Embeddings
should fully
represent all senses of an ambiguous word. 
This criterion becomes more difficult to satisfy as
distributions of senses become more skewed,
but a robust model must be able to overcome this.  

\textbf{Accurate and consistent representation of multifacetedness.}
This criterion addresses
settings with large numbers of facets. 
It is based on the following linguistic
phenomenon, a phenomenon that occurs frequently
crosslinguistically \cite{comrie89typology}.
(i) Words have a large number of facets, including phonetic,
morphological, syntactic, semantic and topical properties.
(ii) Each facet by itself constitutes a small part of
the overall information that a representation
  should capture about a word.

\section{Experimental setup and results}
\seclabel{grammars}
We now design experiments 
to directly evaluate embeddings on the four  criteria.
We proceed as follows. 
First, we design a 
probabilistic context free grammar (PCFG) that
generates a corpus that is a manifestation of the
underlying phenomenon.
Then we train our embedding models
on the corpus.
The embeddings obtained are then evaluated in a classification setting, in which 
we apply a linear SVM \cite{fan08liblinear} to classify embeddings.
Finally, we compare the classification results for different embedding models
and analyze and summarize them. 

\textbf{Selecting embedding models.}
Since
this paper is about developing a new evaluation methodology,
the choice of models is
not
important as long as the models can serve to show that the
proposed methodology reveals interesting
differences with respect to the criteria. 

On the highest level, we can
distinguish two types of distributional
representations. 
\emph{Count vectors}
\cite{sahlgren06word,baroni10distributional,turney10vectorspace} live in a
high-dimensional vector space in which each dimension
roughly corresponds to a (weighted) count of cooccurrence in a
large corpus. \emph{Learned vectors} are learned from large
corpora using machine learning methods: unsupervised methods
such as LSI (e.g.,
\newcite{deerwester90indexing}, \newcite{levy14neural}) and supervised
methods such as neural networks (e.g.,
\newcite{mikolov2013efficient}) and regression (e.g.,
\newcite{pennington14glove}).
Because of the recent popularity of learning-based methods, 
we consider one count-based and five learning-based 
distributional representation models.

The learning-based models are: (i) vLBL (henceforth: LBL)
(vectorized log-bilinear language model)
\cite{MniKav13},
(ii) SkipGram (henceforth: SKIP) (skipgram bag-of-word model),
(iii) CBOW (continuous bag-of-word model \cite{mikolov2013efficient},
(iv) Structured SkipGram (henceforth SSKIP),
\cite{ling15embeddings} and CWindow (henceforth CWIN) (continuous window model) \cite{ling15embeddings}.
These models learn word embeddings for input and target spaces
using neural network models.  

For a  given context, represented by the input space
representations of the left and right neighbors $\vec{v}_{i-1}$
and $\vec{v}_{i+1}$, 
LBL, CBOW and CWIN predict the target space $\vec{v}_{i}$ by combining the contexts.
LBL combines $\vec{v}_{i-1}$ and $\vec{v}_{i+1}$ linearly with position dependent weights and CBOW (resp.\ CWIN) combines them by adding (resp.\ concatenation).
SKIP and SSKIP predict the context words $v_{i-1}$ or $v_{i+1}$ given 
the input space $\vec{v}_{i}$.
For SSKIP, context words are in different spaces depending on their  
position to the input word.
In summary, CBOW and SKIP are learning embeddings using bag-of-word (BoW) models, 
but the other three, CWIN, SSKIP and LBL, are using position dependent models.
We use \texttt{word2vec}\footnote{\url{code.google.com/archive/p/word2vec}} for SKIP and
CBOW, \texttt{wang2vec}\footnote{\url{github.com/wlin12/wang2vec}} for SSKIP and CWIN, and 
\newcite{lai2015compoare}'s implementation\footnote{\url{github.com/licstar/compare}} for LBL.


The count-based model is position-sensitive PPMI, 
\newcite{levy14linguistic}'s 
explicit vector space representation model.\footnote{\url{bitbucket.org/omerlevy/hyperwords}}
For a vocabulary of size $V$, the representation $\vec{w}$ of $w$ is a
vector of size $4V$, consisting of four parts corresponding
to the relative positions $r \in \{-2, -1, 1, 2\}$ with
respect to occurrences of $w$ in the corpus. The entry for
dimension word $v$ in the part of $\vec{w}$ corresponding to
relative position $r$ is the PPMI (positive pointwise
mutual information) weight of $w$ and $v$ for that relative
position. The four parts of the vector are length
normalized.
In this paper, we use only two relative positions: 
$r \in \{-1, 1\}$, so each 
$\vec{w}$ has two parts, corresponding to immediate left and
right neighbors.

\begin{figure}
\small{
\begin{tabular}{rllll}
1&$P(aVb|S)$&=& 1/4 &\\
2&$P(bVa|S)$&=&1/4 &\\\hline
3&$P(aWa|S)$&=& 1/8&\\
4&$P(aWb|S)$&=& 1/8&\\
5&$P(bWa|S)$&=& 1/8&\\
6&$P(bWb|S)$&=& 1/8&\\\hline
7&$P(v_i|V)$&=& 1/5 & $0 \leq i \leq 4$\\
8&$P(w_i|W)$&=& 1/5 & $0 \leq i \leq 4$
\end{tabular}
}
\caption{Global conflation grammar. Words $v_i$ occur
  in a subset of the contexts of words $w_i$, but the global
  count vector signatures are the same.}\figlabel{conflation}
\end{figure}

\subsection{Nonconflation}

\textbf{Grammar.}
The PCFG grammar shown in
\figref{conflation} 
generates $v_i$ words that occur in two types of contexts:
a-b (line 1) and b-a (line 2); and $w_i$
words that also occur in these two contexts (lines 4 and 5),
but in addition occur in a-a (line 3) and b-b
(line 6) contexts. As a result, 
the set of contexts in which
$v_i$ and $w_i$ occur is different, but if we simply count
the number of occurrences in the contexts,
then $v_i$
and $w_i$ cannot be distinguished.

\textbf{Dataset.}
We generated a corpus
of 100,000 sentences.
Words that can occur in a-a and b-b contexts constitute the positive
class, all other words the negative class.
The words $v_3,
v_4, w_3, w_4$ were assigned to the test set, all other words
to the training set.

\textbf{Results.} 
We learn representations of words by our six models and
train one SVM per model; it takes a word representation as
input and outputs +1 (word can occur in a-a/b-b) or -1 (it cannot).
The SVMs trained on PPMI and CBOW representations assigned all four
test set words to the negative class; in particular, $w_3$
and $w_4$ were incorrectly classified. 
Thus, the accuracy of classification for these models
(50\%) was
not better than random.
The SVMs trained on LBL, SSKIP, SSKIP and CWIN representations assigned all four
test set words to the correct class: $v_3$ and $v_4$ were
assigned to the negative class and $w_3$ and $w_4$ were
assigned to the positive class.

\textbf{Discussion.}
The property of embedding models that is relevant here is that
PPMI is an \emph{aggregation model},
which means it 
calculates aggregate statistics for each word and then
computes the final word embedding from these aggregate
statistics. In contrast,
all our learning-based models are \emph{iterative models}:
they iterate over the corpus and each
local context of a word is used as a training instance for
learning its embedding. 

For iterative models, it is common to use
composition of words in the context, as in LBL, CBOW and CWIN. 
Non-compositional iterative models like SKIP and SSKIP  are also popular.
Aggregation models can also use composite features from
context words, but these features are too sparse to be useful.
The reason that the model of
\newcite{agirre09similarity} is rarely used is precisely its
inability to deal with sparseness. 
All widely used
distributional models employ individual word occurrences as 
basic features. 

The bad PPMI results are explained by the fact that it is
an aggregation
model: the  PPMI model cannot distinguish two words with the same global statistics -- as is the case for,
say, $v_3$ and $w_3$.
The bad result of CBOW is probably connected to its weak (addition)
composition of context, although it is an iterative compositional model.
Simple representation of context words with iterative updating
(through backpropagation in each training instance),
can influence the embeddings in a way that SKIP and SSKIP
get good results, although they are 
non-compositional.

As an example of conflation occurring in the English
Wikipedia, consider this simple example. 
We replace all single digits by ``7'' in tokenization. 
We learn PPMI embeddings for the tokens and see that among the one hundred nearest neighbors of ``7'' are the days of the
week, e.g., ``Friday''.
As an example of a conflated feature consider the
word ``falls'' occurring immediately to the right of the
target word. 
The weekdays as well as single
digits often have the immediate right neighbor ``falls'' in
contexts like ``Friday falls on a public holiday'' and ``2
out of 3 falls match'' -- tokenized as ``7 out of 7 falls
match'' -- in World Wrestling Entertainment (WWE).
The left contexts of ``Friday'' and ``7'' are different in
these contexts, but the PPMI model does not record this
information in a way that would make the link to ``falls'' clear.

\begin{figure}
\small{
\begin{tabular}{rllll}
1&$P(AVB|S)$&=&1/2 &\\
2&$P(CWD|S)$&=&1/2 &\\\hline
3&$P(a_i|A)$&=&1/10 & $0 \leq i \leq 9$\\
4&$P(b_i|B)$&=&1/10 & $0 \leq i \leq 9$\\
5&$P(c_i|C)$&=&1/10 & $0 \leq i \leq 9$\\
6&$P(d_i|D)$&=&1/10 & $0 \leq i \leq 9$\\
7&$P(v_i|V)$&=&1/10 & $0 \leq i \leq 9$\\
8&$P(w_i|W)$&=&1/10 & $0 \leq i \leq 9$\\\hline\hline
9&\multicolumn{4}{l}{$L' =  \phantom{ \cup\cup } L(S)$}\\
10&\multicolumn{4}{l}{$\phantom{L' = } \cup \{ a_iu_ib_i |  0 \leq i \leq 9 \}$}\\
11&\multicolumn{4}{l}{$\phantom{L' = } \cup \{ c_ix_id_i |  0 \leq i \leq 9 \}$}                 
\end{tabular}
}
\caption{In language $L'$,  frequent $v_i$ and rare $u_i$
  occur in a-b contexts; frequent $w_i$ and rare $x_i$ occur
  in c-d contexts. Word representations should
  encode possible contexts (a-b vs.\ c-d) for both frequent
  and rare words.}\figlabel{sparseness}
\end{figure}

\subsection{Robustness against sparseness}

\textbf{Grammar.}
The grammar shown in \figref{sparseness} generates frequent
$v_i$ and rare $u_i$ in a-b contexts (lines 1 and 9);
and frequent $w_i$ and rare $x_i$ in c-d contexts (lines 2
and 10).  The language
generated by the PCFG on lines 1--8 is merged on lines 9--11 with the ten contexts
$a_0u_0b_0$ 
\ldots\
$a_9u_9b_9$ (line 9)
and the ten contexts 
$c_0x_0d_0$
\ldots\
$c_9x_9d_9$ (line 10); that is, each
of the $u_i$ and $x_i$ occurs exactly once in the merged
language $L'$, thus modeling the phenomenon of sparseness.

\textbf{Dataset.}
We generated a corpus of 100,000 sentences using the PCFG
(lines 1--8) and added the 20 rare sentences (lines 9--11). 
We label all
words that can occur in c-d  contexts as positive
and all other words as negative.
The singleton words 
$u_i$ and $x_i$
were assigned to the test set, all other words
to the training set.

\textbf{Results.} 
After learning embeddings with different models, 
the SVM trained on PPMI representations assigned all twenty
test words to the negative class. 
This is the correct
decision for the ten $u_i$ (since they cannot occur in a c-d
context), but the incorrect decision for the $x_i$ (since
they can occur in a c-d context).
Thus, the accuracy of classification was 50\% and
not better than random.
The SVMs trained on learning-based representations classified all twenty
test words correctly. 

\textbf{Discussion.}
Representations of rare words in the PPMI model are
sparse. The PPMI representations of the $u_i$ and $x_i$
only contain two nonzero entries,
one entry for an $a_i$ or $c_i$ (left context) and one entry
for a $b_i$ or $d_i$ (right context). Given this sparseness,
it is not surprising that representations are not a
good basis for generalization and PPMI
accuracy is random.

In contrast, learning-based models learn that the $a_i$, $b_i$,
$c_i$ and $d_i$ form four different distributional classes. The final
embeddings of the $a_i$ after learning is
completed are all close to each other and the same is true
for the other three classes. Once the similarity of two
words in the same distributional class (say, the similarity
of $a_5$ and $a_7$) has been learned, the contexts for the
$u_i$ (resp.\ $x_i$) look essentially the same to embedding models as the
contexts of the $v_i$ (resp.\ $w_i$). Thus, the
embeddings learned for 
the $u_i$ will be similar to those learned for the $v_i$.
This explains why learning-based representations achieve perfect
classification accuracy.

This sparseness experiment highlights an important
difference between count vectors
and learned vectors.
Count vector models are less robust
in the face of sparseness and noise because they base their
representations on individual contexts; the overall corpus
distribution is only weakly taken into account, by way of
PPMI weighting. In contrast, learned vector models make much
better use of the overall corpus distribution and they can
leverage second-order effects for learning improved
representations. In our example, the second order effect is
that the model first learns representations for the $a_i$,
$b_i$, $c_i$ and $d_i$  and then uses these as a basis for inferring
the similarity of $u_i$ to $v_i$ and of $x_i$ to $w_i$.

\begin{figure}
\small{
\begin{tabular}{rl@{\hspace{0.03cm}}l@{\hspace{0.03cm}}ll}
1&$P(AV_1B|S)$&=&10/20 &\\\hline
2&$P(CW_1D|S)$&=&9/20 &\\
3&$P(CW_2D|S)$&=&$\beta\cdot$1/20 &\\
4&$P(AW_2B|S)$&=&$(1-\beta)\cdot$1/20 &\\\hline\hline
5&$P(a_i|A)$&=&1/10 & $0 \leq i \leq 9$\\
6&$P(b_i|B)$&=&1/10 & $0 \leq i \leq 9$\\
7&$P(c_i|C)$&=&1/10 & $0 \leq i \leq 9$\\
8&$P(d_i|D)$&=&1/10 & $0 \leq i \leq 9$\\\hline
9&$P(v_{i}|V_1)$&=&1/50 & $0 \leq i \leq 49$\\
10&$P(w_{i}|W_1)$&=&1/45 & $5 \leq i \leq 49$\\
11&$P(w_{i}|W_2)$&=&1/5 & $0 \leq i \leq 4$\\
\end{tabular}
}
\caption{Ambiguity grammar.
$v_i$ and
$w_5 \ldots w_{49}$ 
occur
in a-b and c-d contexts only, respectively. 
$w_0 \ldots w_{4}$ 
are
ambiguous and occur in both contexts. 
}\figlabel{ambiguity}
\end{figure}

\subsection{Robustness against ambiguity}
\seclabel{ambiguity}
\textbf{Grammar.}
The grammar in \figref{ambiguity} generates two types of
contexts that we interpret as two different meanings: 
a-b contexts (lines 1,4) and c-d contexts
(lines 2, 3). $v_i$ occur only in a-b
contexts (line 1),  $w_5 \ldots w_{49}$ occur only
in c-d contexts (line 2); thus, 
they are unambiguous. 
$w_0 \ldots w_{4}$
are ambiguous and occur with
probability $\beta$ in c-d contexts (line 3) 
and with probability ($1-\beta$) in a-b contexts (lines 3, 4). 
The parameter $\beta$ controls the skewedness of the sense
distribution; e.g.,
the two senses are equiprobable for
$\beta=0.5$ and the second sense (line 4) is three
times as probable as the first sense (line 3) for $\beta=0.25$.

\textbf{Dataset.}
The grammar specified in \figref{ambiguity} was used
to generate a training corpus of 100,000 sentences.
Label criterion: 
A word is labeled positive if it can occur in a c-d context, as negative otherwise.
The test set consists of the five
ambiguous words $w_0 \ldots w_4$. 
All other words are assigned to the training set.

Linear SVMs were trained for the 
binary classification task 
on the train set. 
50 trials of this experiment were run for each of eleven
values of $\beta$:
$\beta = 2^{-\alpha}$ where 
$\alpha \in \{ 1.0, 1.1, 1.2, \ldots, 2.0 \}$.
Thus, for the smallest value of $\alpha$, $\alpha=1.0$, the
two senses have the same frequency; for the largest value of
$\alpha$, $\alpha=2.0$, the dominant sense is three times as
frequent as the less frequent sense.
\begin{figure}[tbhp]
\includegraphics[width=0.45\textwidth, height=140px]{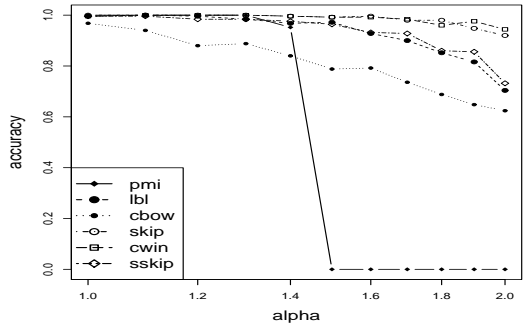}
\caption{
SVM classification results for the ambiguity dataset.
 X-axis: $\alpha = -\log_2 \beta$.
 Y-axis: classification accuracy: 
}
\figlabel{ambiguityeval}
\end{figure}

\textbf{Results.}
\figref{ambiguityeval} shows 
accuracy of the classification on the test set:
the proportion of correctly classified words
out of a total  of 250
(five words each in 50 trials).  

All models perform well for balanced sense frequencies;
e.g., for $\alpha=1.0, \beta=0.5$, the SVMs were all close to 100\% accurate
in predicting that the $w_i$ can occur in a c-d context.
PPMI accuracy 
falls steeply when  $\alpha$ is increased from 1.4 to 1.5.
It has a 100\% error rate for $\alpha \geq 1.5$.
Learning-based models perform better in the order
CBOW (least robust), LBL, SSKIP, SKIP, CWIN (most robust). 
Even for $\alpha=2.0$,
CWIN and SKIP are still close to 100\% accurate.

\textbf{Discussion.}
The evaluation criterion we have used here
is a classification task. The classifier attempts 
to answer  a question that may occur in
an application -- can this word be used in this context? 
Thus, the evaluation criterion is: does the word representation
contain a specific type of information that is needed for
the application.

Another approach to ambiguity is 
to compute multiple representations for a word, one for each sense.
We generally do not yet know what the sense of a word is when
we want to use its word representation, so data-driven
approaches like clustering have been used to create
representations for different usage clusters of words that
may capture some of its senses. For example, 
\newcite{reisinger2010multi} and
\newcite{huang12improving}
cluster the contexts of each word and then
learn a different representation for each cluster.
The main motivation for this approach is the assumption that
single-word distributional representations cannot represent all
senses of a word well \cite{huang12improving}.
However, \newcite{LiJ15a} show that simply increasing the dimensionality 
of single-representation gets comparable results to using multiple-representation.
Our results confirm that a single embedding can be
robust against ambiguity, but also show the main challenge:
skewness of sense distribution.

\def\mytablinesep{0.02cm}

\subsection{Accurate and consistent representation of multifacetedness}
\seclabel{multi}
\begin{figure}
\small{
\begin{tabular}{rl@{\hspace{0.03cm}}l@{\hspace{0.03cm}}ll}
1&$P(NF_n|S)$&=&1/4 &\\[\mytablinesep]
2&$P(AF_a|S)$&=&1/4 &\\[\mytablinesep]
3&$P(NM_n|S)$&=&1/4 &\\[\mytablinesep]
4&$P(AM_f|S)$&=&1/4 &\\\hline\hline
5&$P(n_{i}|N)$&=&1/5 & $0 \leq i \leq 4$\\[\mytablinesep]
6&$P(a_{i}|A)$&=&1/5 & $0 \leq i \leq 4$\\\hline\hline
7&$P(x\uprmdn{nf}{i} U\uprmdn{nf}{i}|F_n)$&=&1/5 &$0 \leq i \leq 4$\\[\mytablinesep]
8&$P(f |U\uprmdn{nf}{i})$&=&1/2 \\[\mytablinesep]
9&$P(\mu(U\uprmdn{nf}{i}) |U\uprmdn{nf}{i})$&=&1/2\\\hline
10&$P(x\uprmdn{af}{i} U\uprmdn{af}{i}|F_a)$&=&1/5 &$0 \leq i \leq 4$\\[\mytablinesep]
11&$P(f |U\uprmdn{af}{i})$&=&1/2 \\[\mytablinesep]
12&$P(\mu(U\uprmdn{af}{i}) |U\uprmdn{af}{i})$&=&1/2\\\hline
13&$P(x\uprmdn{nm}{i} U\uprmdn{nm}{i}|M_n)$&=&1/5 &$0 \leq i \leq 4$\\[\mytablinesep]
14&$P(m |U\uprmdn{nm}{i})$&=&1/2 \\[\mytablinesep]
15&$P(\mu(U\uprmdn{nm}{i}) |U\uprmdn{nm}{i})$&=&1/2\\\hline
16&$P(x\uprmdn{am}{i} U\uprmdn{am}{i}|M_f)$&=&1/5 &$0 \leq i \leq 4$\\[\mytablinesep]
17&$P(m |U\uprmdn{am}{i})$&=&1/2 \\[\mytablinesep]
18&$P(\mu(U\uprmdn{am}{i}) |U\uprmdn{am}{i})$&=&1/2
\end{tabular}
}
\caption{This grammar generates nouns ($x\uprmdn{n.}{i}$) and
  adjectives ($x\uprmdn{a.}{i}$) with masculine ($x\uprmdn{.m}{i}$)
  and feminine ($x\uprmdn{.f}{i}$) gender as well as paradigm features $u_i$. 
  $\mu$ maps each $U$ to one of $\{u_0
  \ldots u_4\}$. $\mu$ is randomly initialized 
and then kept fixed.}\figlabel{similarity}
\end{figure}

\textbf{Grammar.}
The grammar
shown in \figref{similarity} models two syntactic
categories, nouns and adjectives, whose left context is
highly predictable: it is one of five left context
words $n_i$ (resp.\ $a_i$) for nouns, see lines 1, 3, 5
(resp.\ for adjectives, see lines 2, 4, 6).
There are
two grammatical genders: 
feminine (corresponding to the two symbols $F_n$ and $F_a$)
and 
masculine (corresponding to the two symbols $M_n$ and $M_a$). The four
combinations of syntactic category and gender are equally
probable (lines 1--4). 
In addition to
\emph{gender}, nouns and adjectives are distinguished with respect to
\emph{morphological
paradigm}. Line 7 generates one of five feminine nouns
($x\uprmdn{nf}{i}$)
and the corresponding paradigm marker  $U\uprmdn{nf}{i}$. A
noun has two equally probable right contexts: a context
indicating its gender (line 8) and a context indicating its
paradigm (line 9). 
$\mu$ is a function that maps each $U$ to one of 
five morphological paradigms
$\{u_0
\ldots u_4\}$. 
$\mu$ is randomly initialized before a
corpus is generated and kept fixed. 

The function $\mu$
models the assignment of paradigms to nouns and
adjectives. Nouns and adjectives can have different (or
the same) paradigms, but for a given noun or
adjective the paradigm is fixed
and does not change. Lines 7--9 generate gender 
and paradigm markers for feminine nouns, for which we use
the symbols
$x\uprmdn{nf}{i}$. Lines 10--18
cover the three other cases: masculine nouns 
($x\uprmdn{nm}{i}$),
feminine adjectives
($x\uprmdn{af}{i}$)
and masculine adjectives ($x\uprmdn{am}{i}$).

\textbf{Dataset.}
We perform 10 trials. In each trial,
$\mu$ is initialized randomly and
a corpus
of 100,000 sentences is generated.
The train set consists of the  feminine
nouns
($x\uprmdn{nf}{i}$,
line 7)
and
the masculine nouns ($x\uprmdn{nm}{i}$,
 line 13). The test set consists of
the feminine
($x\uprmdn{af}{i}$) and masculine ($x\uprmdn{am}{i}$) adjectives.

\textbf{Results.}
Embeddings have been learned,
SVMs are trained on 
the binary classification task feminine vs.\ masculine and
evaluated on test.
There was not a single
error: accuracy of classifications is 100\% for all embedding models.


\textbf{Discussion.}
The facet gender
is indicated directly by the distribution and easy to learn. For a
noun or adjective $x$, we simply have to check whether $f$
or $m$ occurs to its right anywhere in the corpus. 
PPMI stores this information in two dimensions of
the vectors and the SVM learns this fact perfectly.
The encoding of ``\emph{f} or
$m$ occurs to the right'' is less direct in the learning-based
representation of $x$, but the experiment
demonstrates that they  also reliably encode it and
the SVM reliably picks it up.

It would be possible to encode the facet in just one
bit in a manually designed representation. While all
 representations are less
compact than a one-bit representation -- PPMI uses two real
dimensions, learning-based models use an activation pattern over several
dimensions -- it is still true that most of the capacity of
the embeddings is used for encoding
facets other than gender:
syntactic categories  and paradigms.
Note
that there are five different instances each of
feminine/masculine adjectives,
feminine/masculine nouns and $u_i$ words, but only two gender indicators: $f$
and $m$.  This is a typical scenario across languages:
words are distinguished on a large number of
morphological, grammatical, semantic and other dimensions and each of these
dimensions corresponds to a small fraction of the overall
knowledge we have about a given word.


Point-based tests do not directly evaluate specific facets of words.
In similarity datasets, there is no individual test on
facets -- only full-space similarity is considered. 
There are test cases in analogy that hypothetically evaluate specific facets like gender 
of words, as in king-man+woman=queen.
However, it does not consider the impact of other facets and assumes the only 
difference of ``king'' and ``queen'' is  gender. 
A clear example that words usually differ on many facets,
not just one, is the analogy:
London:England $\sim$ Ankara:Turkey.
\emph{political-capital-of} 
applies to both, 
\emph{cultural-capital-of}  only to London:England since Istanbul is the
cultural capital of Turkey.

To make our argument more clear, we designed an
additional experiment that tries to evaluate gender 
in our dataset based on similarity and analogy
methods. 
In the \emph{similarity evaluation}, we search for the
nearest neighbor of each word and 
accuracy is the proportion of nearest neighbors that have the
same gender as the search word.
In the \emph{analogy evaluation}, we randomly select triples
of the form
$<$$x \uprmmath{c_1g_1}{i}$,$x \uprmmath{c_1g_2}{j}$,$x \uprmmath{c_2g_2}{k}$$>$
where $(c_1,c_2) \in \{ (\textrm{noun}, \textrm{adjective}),
(\textrm{adjective}, \textrm{noun}) \}$ 
and $(g_1,g_2) \in 
\{ ($masculine, feminine$) ,
($feminine, masculine) $\}$. 
We then compute
$\vec{s} =
\vec{x} \uprmmath{c_1g_1}{i}-\vec{x}
\uprmmath{c_1g_2}{j}+\vec{x} \uprmmath{c_2g_2}{k}$
and identify the word whose vector is
closest to $\vec{s}$ where the three vectors
$\vec{x} \uprmmath{c_1g_1}{i}$,
$\vec{x} \uprmmath{c_1g_2}{j}$, $\vec{x} \uprmmath{c_2g_2}{k}$
are excluded. If the nearest neighbor of
$\vec{s}$ is of type $\vec{x} \uprmmath{c_2g_1}{l}$, then the
search is successful; e.g., for $\vec{s} =
\vec{x}\uprmdn{nf}{i}-\vec{x} \uprmdn{nm}{j}+\vec{x} \uprmdn{am}{k}$,
the search is successful if the nearest neighbor is
feminine.
We did this evaluation on the same test set
for PPMI and LBL embedding models. Error rates were 
29\% for PPMI and 25\% for LBL (similarity) and 
16\% for PPMI
and 14\% for  LBL (analogy). 
This high error,
compared to 0\% error for SVM classification, 
indicates it is not possible
to determine the presence of a low
entropy facet accurately and consistently when full-space
similarity and analogy are used as test criteria.

\section{Analysis}
In this section, we first summarize and analyze the
lessons we learned through experiments in \secref{grammars}.
After that, we show how these lessons are supported by 
a real natural-language corpus.

\subsection{Learned lessons}
(i) Two words with clearly different context distributions
should receive different representations. 
Aggregation models fail to do so by calculating global statistics.

(ii) Embedding learning can have different
effectiveness for sparse vs.\ non-sparse events.
Thus, models of representations
should be evaluated with respect to their ability to deal
with sparseness; evaluation data sets should
include rare as well as frequent words.

(iii) Our results in \secref{ambiguity} suggest that single-representation approaches 
can indeed represent different senses of a word.
We did a classification task that roughly
corresponds to the question: does this word have a
particular meaning?
A representation can fail on similarity judgement
computations because less frequent senses occupy a small
part of the capacity of the representation and therefore
have little impact on full-space similarity values. 
Such a failure does not necessarily mean that a particular sense is
not present in the representation and it does not
necessarily mean that single-representation approaches
perform poor on real-world tasks.
However, we saw that even though single-representations do well on
balanced senses, they can pose a challenge for ambiguous
words with skewed senses.


(iv) Lexical information is complex and multifaceted. 
In point-based tests, all dimensions are considered together
and their ability to evaluate specific facets or properties of a word is limited.
The full-space similarity of a word may be highest to a word
that has a different value on a low-entropy facet.
Any good or bad result on these tasks is not sufficient to conclude that the
representation is weak.
The valid criterion of quality is whether information about
the facet is consistently and accurately stored. 

\begin{table}[tbp]
\begin{center}
{\footnotesize
\begin{tabular}{r|cc|cc|cc}
& \multicolumn{2}{|c|}{all entities} &
  \multicolumn{2}{|c|}{head entities} &
  \multicolumn{2}{|c}{tail entities}\\
& MLP & 1NN 
& MLP & 1NN 
& MLP & 1NN\\ 
\hline %

PPMI & 61.6 & 44.0 & 69.2 & 63.8 & 43.0 & 28.5 \\ 
LBL  & 63.5 & 51.7 & 72.7 & 66.4 & 44.1 & 32.8 \\ 
CBOW & 63.0 & 53.5 & 71.7 & 69.4 & 39.1 & 29.9  \\ 
CWIN & 66.1 & 53.0 & 73.5 & 68.6 & 46.8 & 31.4\\ 
SKIP & 64.5 & \textbf{57.1 }& 69.9 & \textbf{71.5} & \textbf{49.8} & \textbf{34.0}\\ 
SSKIP& \textbf{66.2} & 52.8 & \textbf{73.9} & 68.5 & 45.5 & 31.4
\end{tabular} 
}
\end{center}
\caption{Entity typing results using embeddings learned with
different models.}
\tablabel{entity}
\end{table}

\subsection{Extrinsic evaluation: entity typing}
To support the case for sub-space evaluation 
and also to introduce a new extrinsic task that uses the embeddings
directly in supervised classification,  
we address a \emph{fine-grained entity typing} task.

Learning taxonomic properties or types of words 
has been used as an evaluation method 
for word embeddings \cite{Rubinstein15embed}.
Since available word typing datasets are quite small 
(cf. \newcite{baroni14predict}, \newcite{Rubinstein15embed}), 
entity typing can be a promising alternative,
which enables to do 
supervised classification instead of unsupervised clustering.
Entities, like other words, have many properties
and therefore belong to several semantic types, e.g., 
``Barack Obama'' is a \textsc{politician}, \textsc{author} and 
\textsc{award\_winner}.
We perform entity typing by
learning types of knowledge base entities from their
embeddings; this requires
looking at sub-spaces because each entity can 
belong to multiple types.

We adopt the setup of
\newcite{figment15} who present a dataset of Freebase entities;\footnote{\url{cistern.cis.lmu.de/figment}}
there are 102 types (e.g.,
\textsc{politician} \textsc{food},
\textsc{location-cemetery}) and most entities have several.
More specifically, we use a multi-layer-perceptron (MLP) with one hidden layer
to classify entity embeddings to 102 FIGER types.
To show the limit of point-based evaluation, 
we also  experimentally test an entity typing model based 
on \emph{cosine similarity} of entity embeddings.
To each test entity, we assign all types of the
entity closest to it in the train set.
We call this approach 1NN (kNN for $k=1$).\footnote{We tried 
other values of $k$, but results were not better.}

We take part of ClueWeb, which is annotated with Freebase entities using automatic annotation of FACC1\footnote{{\url{lemurproject.org/clueweb12/FACC1}}}
\cite{gabrilovich2013facc1}, as our corpus.
We then replace all mentions of entities with their Freebase identifier
and learn embeddings of words and entities in the same space.
Our corpus has around 6 million sentences with at least one annotated entity.
We calculate embeddings using our different models. 
Our hyperparameters: 
for learning-based models: dim=100, neg=10, iterations=20, window=1,
sub=$10^{-3}$;
for PPMI: SVD-dim=100, neg=1, window=1, cds=0.75, sub=$10^{-3}$, eig=0.5.
See \cite{levy15improve} for more information about the meaning of
hyperparameters.

\tabref{entity} gives results on test
for all  (about 60,000 entities), 
head (freq $>$
100; about 12,200 entities) and tail (freq $<$ 5; about 10,000 entities).
The MLP models consistently outperform 1NN on all and tail
entities.  This supports our hypothesis that only part of
the information about types that is present in the vectors
can be determined by similarity-based methods that use the
overall direction of vectors, i.e., full-space similarity.

There is little correlation between results of MLP and 1NN
in all and head entities, and the correlation between their
results in tail entities is high.\footnote{The spearman
  correlation between MLP and 1NN for all=0.31, head=0.03,
  tail=0.75.}  For example, for all entities, using 1NN,
SKIP is 4.3\% (4.1\%) better, and using MLP is 1.7\% (1.6\%)
worse than SSKIP (CWIN).  The good performance of SKIP on
1NN using cosine similarity can be related to its objective
function, which maximizes the cosine similarity of
cooccuring token embeddings.


The important question is not similarity, but whether the
information about a specific 
type exists in the entity embeddings or not. Our results
confirm our previous observation that a classification 
by looking at subspaces is needed 
to answer this question. In contrast, based on
full-space similarity, one can infer little
about the quality of embeddings.
Based on our results, SSKIP and CWIN embeddings contain more accurate
and consistent information because MLP classifier gives
better results for them.
However, if we considered 1NN for comparison, SKIP and CBOW
would be superior.

\section{Conclusion and future work}
We have introduced a new way of evaluating
distributional representation models.  
As an alternative to the
common evaluation tasks, we proposed to identify
generic criteria that are important for an embedding model 
to represent properties of words accurately and consistently.  
We
suggested four criteria based on fundamental characteristics
of natural language and designed tests that 
evaluate models on the criteria.
We developed this evaluation
methodology using PCFG-generated corpora and applied it on a
case study to compare different models of learning distributional  representations.


While we showed important differences of the embedding
models, the goal was not to do a comprehensive comparison of
them.  We proposed an innovative way of doing intrinsic
evaluation of embeddings.  Our evaluation method gave direct
insight about the quality of embeddings.  Additionally,
while most intrinsic evaluations consider word vectors as
points, we used classifiers that identify different small
subspaces of the full space.  This is an important
desideratum when designing evaluation methods because of the
multifacetedness of natural language words: they have a
large number of properties, each of which only occupies a
small proportion of the full-space capacity of the
embedding.

Based on this paper,
there are serveral lines of investigation we plan to conduct
in the future. (i)
We will attempt to support our results on
artificially generated corpora by conducting experiments on
\emph{real natural language data}.  (ii) We will study  the
\emph{coverage of our four criteria} in evaluating word
representations. (iii)
We modeled the four criteria using separate PCFGs,
 but they could also be modeled by one single
unified PCFG. One question that arises is then to what
extent the four criteria are orthogonal and to what extent
interdependent.
A single unified grammar may make it harder
to interpret the results, but may give additional and more fine-grained
insights as to how the performance of embedding models is
influenced by different fundamental properties of natural
language and their interactions.

Finally, we have made the simplifying assumption in this
paper that the best conceptual framework for thinking about
embeddings is that the embedding space can be
\emph{decomposed into subspaces}: either into completely
orthogonal subspaces or -- less radically -- into partially
``overlapping'' subspaces.  Furthermore, we have made the
assumption that the smoothness and robustness properties
that are the main reasons why embeddings are used in NLP can
be reduced to \emph{similarities in subspaces}.  See
\newcite{rothe16ultradense} and \newcite{rothe16spectrum}
for work that makes similar assumptions.

The fundamental assumptions here are decomposability and
linearity. 
The smoothness properties could be much more
complicated. However even if this was the case, then much of the general framework of what we have
presented in this paper would still apply; e.g., 
the criterion that a particular facet be fully
and correctly represented  is as important as before. But
the validity of the assumption that
embedding spaces can be decomposed into ``linear'' subspaces
should be investigated in the future.

\textbf{Acknowledgments.} This work was supported by DFG
(SCHU 2246/8-2).


\bibliography{buecher}

\bibliographystyle{acl2016}

\end{document}